\documentclass[10pt,twocolumn,letterpaper]{article}

\usepackage{cvpr}
\usepackage{times}
\usepackage{epsfig}
\usepackage{graphicx}
\usepackage{amsmath}
\usepackage{amssymb}
\usepackage{subfigure}
\usepackage{slashbox}
\usepackage[us,12hr]{datetime}

\usepackage[pagebackref=true,breaklinks=true,letterpaper=true,colorlinks,bookmarks=false]{hyperref}

\cvprfinalcopy 

\def\cvprPaperID{1608} 

\ifcvprfinal\fi
\begin{document}

\title{CRAFT: Complementary Recommendations Using\\Adversarial Feature Transformer}

\author{Cong Phuoc Huynh \qquad	
	Arridhana Ciptadi \qquad
	Ambrish Tyagi\qquad
	Amit Agrawal\\
	Amazon.com\\
{\tt\small \{conghuyn, ambrisht, aaagrawa\}@amazon.com}
}

\maketitle



\begin{abstract}

Traditional approaches for complementary product recommendations rely on behavioral and non-visual data such as customer co-views or co-buys. However, certain domains such as fashion are primarily visual. We propose a framework that harnesses visual cues in an unsupervised manner to learn the distribution of co-occurring complementary items in real world images. Our model learns a non-linear transformation between the two manifolds of source and target complementary item categories (e.g., tops and bottoms in outfits). Given a large dataset of images containing instances of co-occurring object categories, we train a generative transformer network directly on the feature representation space by casting it as an adversarial optimization problem. Such a conditional generative model can produce multiple novel samples of complementary items (in the feature space) for a given query item. The final recommendations are selected from the closest real world examples to the synthesized complementary features. We apply our framework to the task of recommending complementary tops for a given bottom clothing item. The recommendations made by our system are diverse, and are favored by human experts over the baseline approaches.

\end{abstract}


\section{Introduction}
\label{sect:Introduction}

Recommendation algorithms are central to many commercial applications, particularly for online shopping. In domains such as fashion, customers are looking for clothing recommendations that visually complement their current outfits, styles, and wardrobe. 
Traditional content-based and collaborative recommendation algorithms~\cite{Adomavicius2005Recommender,Lew2006CMIR} do not make use of the visual cues to suggest complementary items. Among these, collaborative filtering~\cite{Koren2011Collaborative,Melville2002ContentCF} is a commonly used approach, which primarily relies on behavioral and historical data such as co-purchases, co-views, and past purchases to suggest new items to customers.
In contrast to these approaches, this work addresses the problem of providing complementary item recommendations for a given query item based on visual cues.

\begin{figure}[tbp]
	\centering
	\includegraphics[width=0.48\textwidth,trim={2.5cm .7cm 2.5cm 0.6cm},clip]{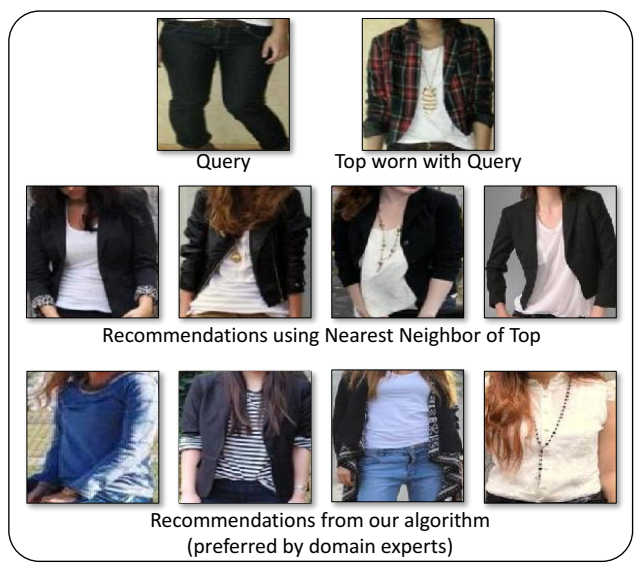}
	\caption{Recommending tops for a given query bottom. Tops that are visually similar to the \textit{actual} top worn with the query item are acceptable options, but lack diversity. Our approach generates both complementary and diverse recommendations that are also preferred by the fashion specialists. Please see the supplementary material for more examples.}
	\label{fig:top_nn_bad}
	\vspace*{-3ex}
\end{figure}

\begin{figure*}
	\centering
	\includegraphics[width=\linewidth,
	trim={0.75cm 10.75cm 0.75cm 5cm},clip]{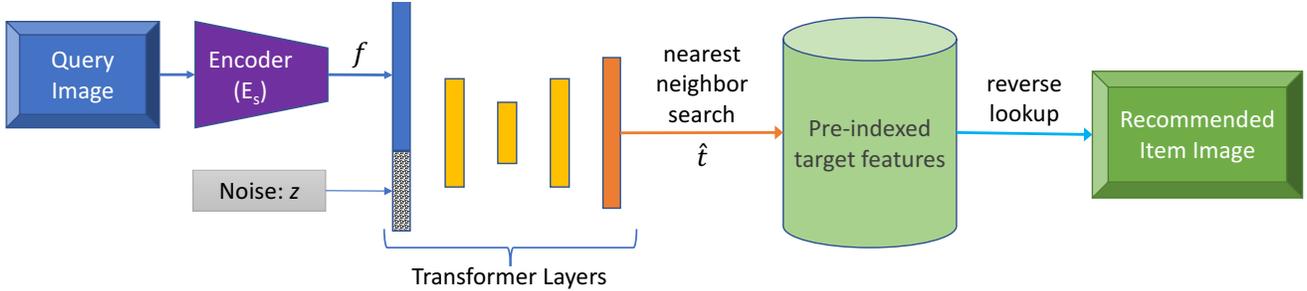}
	\caption[Inference]{Generating recommendations using the proposed CRAFT network.}
	\label{fig:recommendation_workflow}
    \vspace{-2ex}
\end{figure*}

We develop an unsupervised learning approach for Complementary Recommendation using Adversarial Feature Transform (CRAFT), by learning the co-occurrence of item pairs in real images.
The assumption here is that the co-occurrence frequency of item pairs is a strong indicator of the likelihood of their complementary relationship. We define an adversarial process to train a conditional generative transformer network that can learn the joint distribution of item pairs by observing samples from the real distribution, \ie image features of co-occurring items.

In contrast to traditional generative adversarial approaches that aim to directly synthesize images, our generative transformer network is trained on and generates samples in the \emph{feature} space. To generate a recommendation for a given query image, we first extract its features via a pre-trained encoder network, $E_s$ (Figure~\ref{fig:recommendation_workflow}). The query feature, $f$, along with a sampled noise vector, $z$, is fed to the transformer network to generate the feature vector for a complementary item. We then use the generated feature, $\hat{t}$, to retrieve the recommended target image by performing a nearest neighbor search on a pre-indexed candidate subspace. Our proposed approach is general (modeling visual co-occurrences) and can be applied to different domains such as fashion, home design, \etc.

Our approach is a novel and unique way of utilizing generative adversarial training with several advantages over traditional Generative Adversarial Network (GAN)~\cite{GAN} based image generation. While the quality of visual image generation using GANs has improved significantly (especially for faces~\cite{ProgressiveGAN}), it still lacks the realism required for many real-world applications, such as in the fashion/apparel industry. More importantly, the goal of a recommendation system in such applications is often not to generate synthetic images, but rather to recommend real images from a catalog of items. An approach that generates synthetic images will thus still need to perform a search, which is typically done in the feature space, to find the most visually similar images in the catalog. CRAFT directly generates features of the recommended items, thereby bypassing the need to generate synthetic images and enabling a simpler and more efficient algorithm. By working in the feature space, we can use a simpler network architecture, which improves stability during training and avoids common pitfalls such as mode collapse~\cite{ModeRegularizedICLR2017}.

We evaluate our algorithm (CRAFT) on a real-world problem of recommending complementary top clothing items for a given query bottom. For this task, a recommendation algorithm is expected to produce both complementary and diverse set of results. Figure~\ref{fig:top_nn_bad} shows an example of the recommendation results from CRAFT and from a na\"ive approach of recommending nearest neighbors of the top. It is clear that generating recommendations by simply finding visually similar tops lacks the capability of producing \textit{diverse} set of results.

Given that the assessment of fashion recommendations is a subjective problem, we conducted rigorous studies with fashion specialists to evaluate the quality of our algorithm. Through these studies, we demonstrate the effectiveness of our approach as compared to several baseline approaches (Section~\ref{subsect:baseline-algorithms}).

\section{Related Work}
\label{sect:related-work}

\begin{table*}
	\centering
	\small
	\begin{tabular}{|l|c|c|c|}
		\hline
		\multicolumn{1}{|c|}{GAN Input} & Generative & Output & Example  \\
		&  (w/ random seed) &  &\\
		\hline
		N/A & Yes & Image & Image Generation~\cite{GAN} \\
		Image  & No & Image & Image-to-Image Translation~\cite{CycleGANICCV2017} \\
		Image + Attribute  & No & Image & Image Manipulation~\cite{DBLP:journals/corr/LampleZUBDR17} \\
		Synthetic Image & No & Image & Adding Realism~\cite{ShrivastavaCVPR2017} \\
		Synthetic Image & Yes & Image & Adding Realism~\cite{DBLP:journals/corr/BousmalisSDEK16} \\
		Image & No & Features & Domain Adaptation~\cite{ADDATrevorCVPR2017} \\
		Features & Yes & Features & Ours \\
		\hline
	\end{tabular}
	\vspace{0.15cm}
	\label{table:Comp}
	\caption{Similarities and differences between our approach and those that use adversarial loss for training. }
	\vspace*{-3ex}
\end{table*}

\textbf{Generative Adversarial Networks}: GANs~\cite{GAN} have recently emerged as a powerful framework for learning generative models of complex data distributions. They have shown impressive results for various tasks including image generation~\cite{ProgressiveGAN,DBLP:journals/corr/LuTT17}, image-to-image translation~\cite{Isola2017,CycleGANICCV2017}, domain adaptation~\cite{DBLP:journals/corr/BousmalisSDEK16,ShrivastavaCVPR2017,ADDATrevorCVPR2017}, \etc. In a GAN framework, a generator is trained to synthesize samples from a latent distribution and a discriminator network is used to distinguish between synthetic and real samples. The generator's goal is to fool the discriminator by producing samples that are as close to real data as possible. A recent work by Zhu~\etal~\cite{ZhuICCV2017} used the GAN framework to generate new clothing on a wearer. Our approach differs from these methods since we do not aim to generate an image of the complementary item. Instead, we use the adversarial training framework to learn the joint distribution between the source and target \textit{features} in an \textit{unsupervised} manner. We train a transformer that takes as input a random noise vector as well as the features of a query image and generates a feature vector representation of a complementary item.

The GAN paradigm has also found applications in the areas of image manipulation and image transformation
~\cite{DBLP:journals/corr/BousmalisSDEK16,Isola2017,ShrivastavaCVPR2017,CycleGANICCV2017}. For example, Srivastava~\etal~\cite{ShrivastavaCVPR2017} add realism to synthetic data by training an adversarial network that transforms a synthetic image into a real image. While such an approach can be applied to transform a given image into that of a complementary item, it only provides a fixed mapping. In contrast, ours is a generative approach that can provide multiple complementary items by learning the joint distribution in the feature space. Contrary to methods such as CycleGAN~\cite{CycleGANICCV2017} that perform image-to-image translation using raw pixels, our approach works directly in the feature space. Feature-based domain adaptation approaches such as~\cite{ADDATrevorCVPR2017} attempt to directly learn a visual encoder for the target domain by minimizing an adversarial loss defined on the source and target features.
In contrast, we train a generative transformer network that operates in the feature space. Table~\ref{table:Comp} shows similarities and differences between our approach and those that use adversarial loss for training.

\textbf{Unsupervised Learning:} 
Recent applications of unsupervised learning for visual tasks include object discovery in videos~\cite{CroitoruICCV2017}. In addition, there have been demonstrations of self-supervised learning~\cite{DoerschICCV2017} for the tasks of image colorization, image in-painting, hole filling, jigsaw puzzle solving from image patches, future frame prediction using video data~\cite{VondrickCVPR2017}, \etc. In the  fashion domain, annotated data are typically used for predicting fashion-related attributes and matching street-to-catalog images~\cite{WhereToBuyItICCV15,StreetToShop}. These approaches involve visual search to find \textit{similar} looking items, whereas our approach is focused on finding complementary items. Furthermore, our approach is unsupervised: we only take as input a set of images to learn the feature transformation between complementary objects.

\textbf{Recommendation:}
There is a rich body of literature on using behavioral customer data such as browsing and purchasing history to develop recommender systems~\cite{Koren2011Collaborative,DBLP:journals/corr/ZhangYS17aa}.
Specific to the fashion domain, McAuley~\etal~\cite{McATarShiHen15} employed convolution neural network (CNN) features and non-visual data to build a personalized model of user's preference. In~\cite{He2016Compatibility}, the authors proposed a mixture of (non-metric) embeddings to recommend visually compatible items in several categories. A related approach is to learn a common embedding across categories and use a metric function in the embedding subspace as a measure of visual compatibility~\cite{Veit2015Style}. 
In~\cite{HanWJD17}, the authors proposed to learn a bi-directional Long Short Term Memory (LSTM) model in a supervised manner, to suggest items that complement each other in an entire outfit.

The aforementioned recommendation approaches use customer's behavioral data as training labels. Behavioral signals do not necessarily reflect that items viewed or purchased together are visually complementary. In contrast, our unsupervised approach learns item co-occurrences from only visual data. In multiple methods~\cite{HanWJD17,McATarShiHen15,Veit2015Style}, the recommendation model is non-generative in the sense that it can only evaluate the compatibility between two given items. In others~\cite{He2016Compatibility}, the diversity of recommendation is limited by the (fixed) number of embeddings employed. In contrast, our \textit{generative} model is not subject to such a constraint, thanks to its ability to sample an infinite amount of noise vectors.

\section{Generative Feature Transformer Network}
\label{sec:transformer-details}
This section describes our generative approach for complementary recommendation based on the co-occurrence of item pairs in real-world images. This could include, for example, combinations of top and bottom clothing items that people wear as part of their outfits or pairs of furniture items such as sofa and chairs that are present in most household scenes. We hypothesize that learning the joint distribution of such pairs can be useful for recommending new items that complement a given query. We adopt an adversarial learning paradigm, where our transformer network learns to generate features of the complementary items conditioned on the query item.

\begin{figure*}[t]
	\centering
	\includegraphics[width=0.85\linewidth,trim={0.2cm 6cm 0.5cm 5cm},clip]{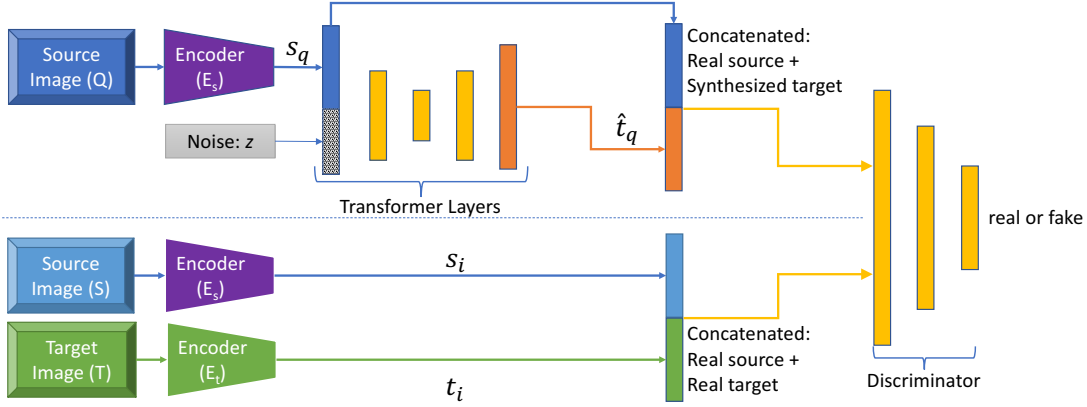}
	\caption[Architecture]{Architecture for the CRAFT framework. The transformer is trained using the adversarial loss to generate the target features conditioned on the source features and a sampled noise vector.}
	\label{fig:architecture}
    \vspace*{-3ex}
\end{figure*}

\subsection{Network Architecture}

In contrast to traditional GANs, our approach synthesizes visual features rather than images, which offers several advantages. It is more tractable to sample the feature space rather than the image space. The bulk of computation in the convolution and transpose convolution layers, which are usually required for feature learning and image synthesis, can be avoided, leading to a simpler architecture and stable training.

We first select an appropriate visual representations for the source and target images. The fixed feature representations (encodings) are generally derived from pre-trained CNNs. Typically, it is advisable to use application-specific feature representations, \eg, apparel feature embeddings for clothing recommendations, but a general representation such as one trained on ImageNet~\cite{imagenet_cvpr09} or MS-COCO~\cite{MS_COCO} offer robust alternatives. Figure~\ref{fig:architecture} depicts the overall architecture of the CRAFT network. The source and the target feature encoders, $E_s$ and $E_t$, respectively, are fixed and are used to generate feature vectors for training and inference.

Our architecture resembles traditional GAN designs with two main components: a conditional feature transformer and a discriminator. The role of the feature transformer is to transform the source feature $s_q$ into a complementary target feature $\hat{t}_q$. The input to the transformer also consists of a random noise vector $z$ sampled uniformly from a unit sphere in a $d_z$-dimensional space. By design, the transformer is generative since it is able to sample various features in the target domain.

As discussed, since our approach works in the feature space, we can adopt a simple architecture for the feature transformer and discriminator. The transformer consists of several fully-connected layers, each followed by batch normalization~\cite{Ioffe2015BN} and leaky ReLU~\cite{Mass2013LReLU} activation layers. The discriminator is commensurate to the transformer in capacity, consisting of the same number of layers. This helps balance the power between the transformer and the discriminator in the two-player game, leading to stable training and convergence.


\subsection{Training}
Our training data consists of $N$ co-occurring feature pairs $\mathcal{C} = \{(s_i, t_i), i=1, \ldots, N\}$, where $s_i \in \mathbb{R}^{d_s}$ and $t_i \in \mathbb{R}^{d_t}$ denote the features corresponding to the source and the target images, respectively.
Given a sample $s_q$ from the source space, the complementary recommendation task is to generate target features $\{\hat{t}_q\}$ that maximizes the likelihood that the pair $(s_q, \hat{t}_q)$ belongs to the joint distribution $p_\mathcal{C}$ represented by the training data. To this end, we model the composition of layers in the feature transformer and the discriminator as two functions  $T_\phi(s,z): (s, z) \mapsto \hat{t}$ and $D_\theta(s,t): (s,t) \mapsto [0,1]$, respectively. Here, $\phi$ and $\theta$ are the learnable parameters of the two players, transformer and discriminator, respectively, and $(s, t)$ is a pair of source and target feature vectors, and $z$ is a random noise vector.

The training process emulates an adversarial game between the feature transformer and the discriminator, where the discriminator aims to classify feature pairs as real (co-occurring) or synthetic. On the other hand, the feature transformer synthesizes target features $\{\hat{t}_q\}$ conditioned on a given source feature $s_q$. Its objective is to fool the discriminator into the belief that $\hat{t}_q$  co-occurs with $s_q$. The feedback from the discriminator encourages the transformer to produce a target feature $\hat{t}_q$ so as to maximize the co-occurrence probability of the synthetic pair.

The adversarial game can be formulated as a mini-max optimization problem. The optimization approach can be implemented by alternating the training of the discriminator and the feature transformer. In the discriminator step (D-step), the discriminator's goal is to assign a binary label, \ie $0$ to the synthesized feature pair $(s_q, \hat{t}_q)$, where $\hat{t}_q=T_\phi(s_q, z)$, and $1$ to an actual pair $(s_i, t_i)$. The discriminator maximizes the cross entropy loss in Equation~\ref{eqn:d_loss}.
\begin{equation}
\begin{split}
\mathcal{L}_D &\triangleq \mathbb{E}_{(s_i, t_i) \sim p_\mathcal{C}} \log D_\theta(s_i, t_i)\\
&+\mathbb{E}_{z \sim p_z, s_q \sim p_s} \log(1-D_\theta(s_q, T_\phi(s_q, z) ))
\label{eqn:d_loss}
\end{split}
\end{equation}
where $p_z$ and $p_s$ are the probability distribution function (pdf) of the random noise and the source feature.

The feature transformer maximizes the likelihood that the discriminator recognizes synthetic pairs as belonging to the data-generating (joint) distribution $\mathcal{C}$, \ie it assigns a label $1$ to such pairs. Therefore, the transformer step (T-step) aims to minimize the loss in Equation~\ref{eqn:t_loss}.
\begin{equation}
\begin{split}
\mathcal{L}_T 
			&= \mathbb{E}_{z \sim p_z, s_q \sim p_s} \log(1-D_\theta(s_q, T_\phi(s_q, z)))
\label{eqn:t_loss}
\end{split}
\end{equation}

The overall objective function of the adversarial training process is formulated in Equation~\ref{eqn:adversarial_loss}.
\begin{equation}
\begin{split}
& \min_{\phi} \max_{\theta} \mathcal{L} \triangleq
\mathbb{E}_{(s_i, t_i) \sim p_\mathcal{C}} \log D_\theta(s_i, t_i)\\
& +\mathbb{E}_{z \sim p_z, s_q \sim p_s} \log(1-D_\theta(s_q, T_\phi(s_q, z)))
\end{split}
\label{eqn:adversarial_loss}
\end{equation}

While our approach is completely unsupervised and does not require labels for complementary relationships, it can be easily extended to train a semi-supervised transformer that can benefit from additional complementary annotations. For example, if we know that certain complementary combinations are better or are ranked higher than others, then we can employ an additional discriminator that can take concatenated source and target features from labeled data and predict the ranking of these pairs.




\subsection{Generating Recommendations}
The recommendation work-flow is depicted in Figure~\ref{fig:recommendation_workflow}. Here, we retain only the transformer's layers shown in Figure~\ref{fig:architecture} for recommendation.
From a query image, the query feature $f$ is extracted by the source encoder, $E_s$, and multiple samples of transformed features $\{\hat{t}_i\}$ are generated by sampling random vectors $\{z_i\}$. This allows us to generate a diverse set of complementary recommendations by sampling the underlying conditional probability distribution function. Subsequently, we perform a nearest neighbor search within a set of pre-indexed target features extracted using the same target encoder, $E_t$, used during training. Actual recommendation images are retrieved by a reverse lookup that maps the selected features to the original target images. 
 

\section{Experimental Setup}
\label{sect:experiments}
We demonstrate the efficacy of our approach by applying it to the problem of complementary apparel recommendations.
Specifically, we train the generative transformer network to synthesize features for top clothing items that are visually compatible to a given query bottom item. The generated features are used to retrieve the images of the nearest neighbors in a pre-indexed catalog of candidate top items as the complementary recommendations.

\subsection{Datasets}
\label{subsect:datasets}
We trained the proposed CRAFT network from scratch on 
unlabeled images, without the need for any human annotation to define complementary relationships. 
Our training data consisted of 473k full-length outfit images, each containing a top and a bottom clothing part, under the assumption that their outfits are highly complementary.

Each image was preprocessed to extract the regions of interest (ROIs) corresponding to the top and bottom clothing items.  
To extract clothing part ROIs, we trained a semantic segmentation network~\cite{Zhao2017pspnet} on the ``Human Parsing in the Wild'' dataset~\cite{Liang2015HPW}. We consolidate the original labels in the dataset into $15$ labels, where top clothing items correspond to the label ``upper-clothes'' and bottom ones correspond to ``pants'' and ``skirt''. Using this segmentation network, we parse the training images into clothing parts, based on which tight bounding boxes around the segments corresponding to top and bottom regions are selected. In this manner, the training pairs are obtained automatically from images without the need for manual complementary annotation. 
Only the pre-processing stage requires supervised training. Having obtained the input feature from this stage, the feature transformer was trained in an unsupervised manner. 

As discussed earlier, we use a fixed visual feature encoding for both the training and the recommendation process. To this end, we extract the global averaging pooling feature encoded by the Inception-v4 model~\cite{Szegedy2016Inception}. Rather than working in the original feature space with $1536$ dimensions, we further reduce the dimensionality to $128$ by Principal Component Analysis. This helps address the curse of dimensionality and further reduces the computational load.

\subsection{Training and Network Parameters}
We use the Adam optimizer~\cite{KingmaB14} with starting learning rate of $0.0002$ for both the discriminator and the transformer networks. To improve training stability, we use one-sided label noise~\cite{salimans2016improved}. Each minibatch for training the discriminator consists of an equal proportion of synthetic and real feature pairs. Our transformer network is composed of $3$ fully connected layers with $256$ channels in the first two layers and $128$ channels in the third.
Our discriminator is composed of $3$ fully connected layers with $256$, $256$, and $1$ channel(s), respectively. The noise vector $z$ is uniformly sampled from the unit sphere in $\mathbb{R}^{128}$. We use leaky ReLU ($\alpha=0.2$) and batch normalization for the first two layers of both the transformer and the discriminator.

\subsection{Baseline Algorithms}
\label{subsect:baseline-algorithms}

To demonstrate the effectiveness of CRAFT, we compare it with the following baseline algorithms:

\textbf{Random recommendations:} A trivial baseline generates random recommendations from a given set of candidate options, referred to as \textit{Random}. A random selection can offer a diverse set of target items, but they may not necessarily be complementary to the query item.

\textbf{Nearest neighbors of source items:} A strong and useful baseline method is to find items similar to the query in the source space, and recommend their \textit{corresponding} target items. For example, if the query item is blue jeans, this approach will recommend various tops that other people have worn with similar blue jeans. Note that these can be retrieved by performing a visual search for the nearest neighbors of the bottom item. We refer to this method as \textit{NN-Source}.


\textbf{Incompatible recommendations:} Additionally, we illustrate that CRAFT has not only learned to recommend complementary apparel items, but also learned the concept of visual incompatibility. We design a method for generating \textit{Incompatible} recommendations by suggesting tops that are assigned low discriminator scores by the CRAFT network.



\section{Results}
\label{sect:subjective-evaluation}

Now we present an in-depth analysis of our algorithm and comparisons with baseline methods. In Section~\ref{subsect:visualze-top-space}, we visualize how the learned transformer network \textit{dynamically} reacts to given queries in terms of assigning compatibility scores for candidate tops. We then show qualitative results of recommendations produced by various algorithms in Section~\ref{subsect:qualitative-results}. Finally, we describe our study with domain experts in Section~\ref{subsect:user-study-design} and analyze the results in Section~\ref{subsect:user-study-analysis}.

\subsection{Visualization of The Discriminator Output}
\label{subsect:visualze-top-space}
\begin{figure}
	\centering
	\includegraphics[width=\linewidth,height=1.65in]{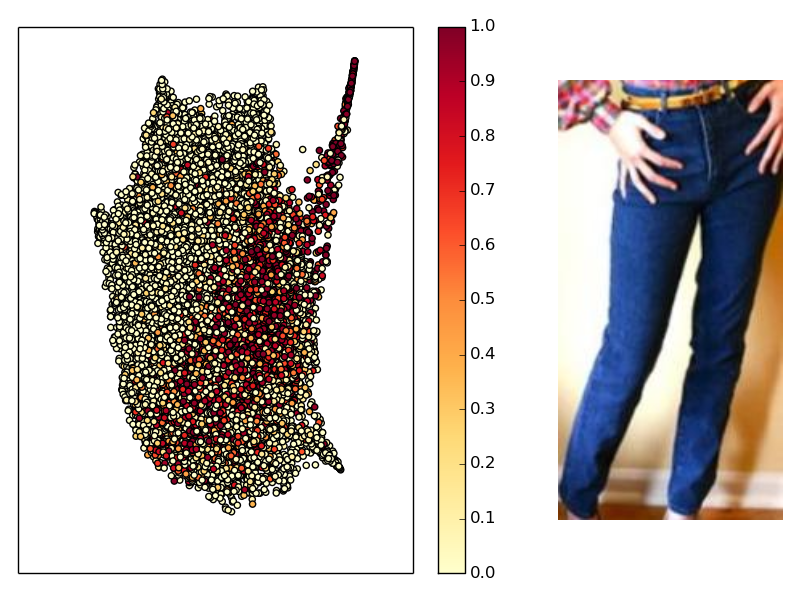}\\
	\includegraphics[width=\linewidth,height=1.65in]{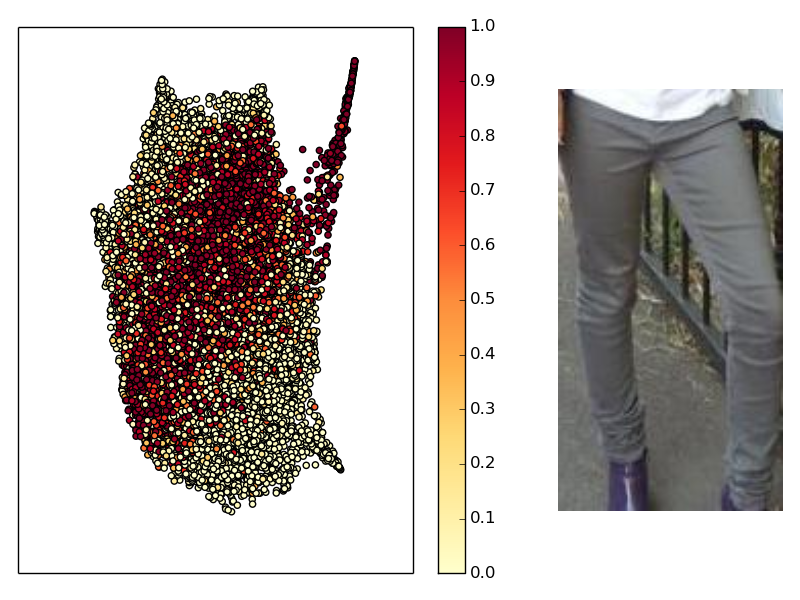}\\
	\includegraphics[width=\linewidth,height=1.65in]{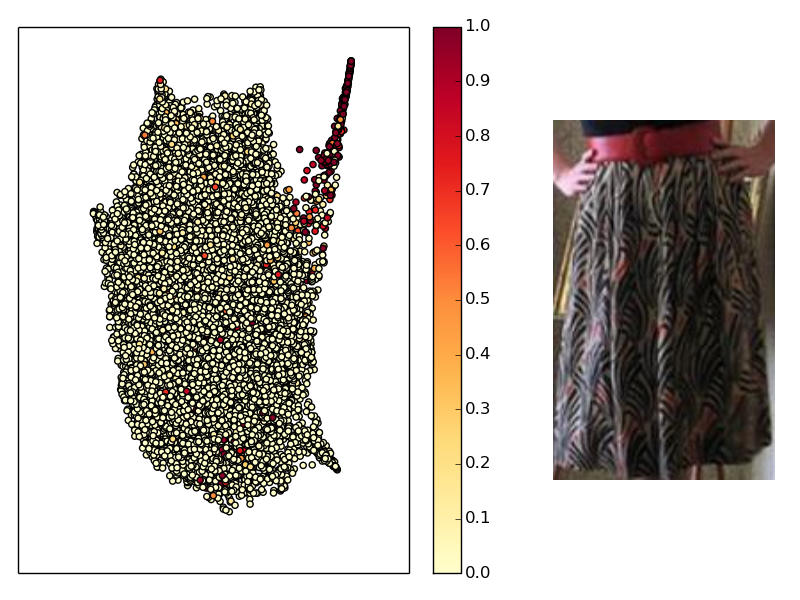}
	\caption{Each row shows a 2D t-SNE embedding of all the candidate tops (left) with the corresponding query image (right). The colors represent the discriminator score for tops conditioned on the query (red: high score, yellow: low score). Note that the discriminator in CRAFT is able to learn that common bottoms such as blue jeans and gray pants are compatible with a wide range of tops as compared to rarer query items such as the patterned skirt shown in the last row.}
	\label{fig:discriminator_scores}
\vspace*{-3ex}
\end{figure}
In order to visualize the space of candidate top items, we projected them to a two-dimensional (2D) subspace using t-SNE~\cite{maaten2008visualizing}. The discriminator output can be seen as a proxy for the compatibility score between any top and a given query item. Each row in Figure~\ref{fig:discriminator_scores} shows 2D embedding of all the tops in the dataset, color coded by the discriminator score for each top, given the bottom query item (shown on the right). Note that the compatibility scores for various candidate tops change depending on the query bottom. The yellow colors in the t-SNE plot denote low compatibility, while shades of orange to red denote high compatibility (see color bar). It is interesting to note how universal items such as blue jeans or gray pants are compatible with a large set of candidate tops, while rare bottoms like the richly textured pattern skirt shown on the bottom row are compatible with only a handful of tops. This illustrates that CRAFT is able to model the distribution of real item pairs.



\subsection{Qualitative Results}
\label{subsect:qualitative-results}
\begin{figure*}[htb]
	\centering
	\subfigure[Complementary recommendation for a common query item (dark jeans)]{
		\includegraphics[width=\textwidth,trim={0.2cm 0.3cm 0.2cm 0.3cm},clip]{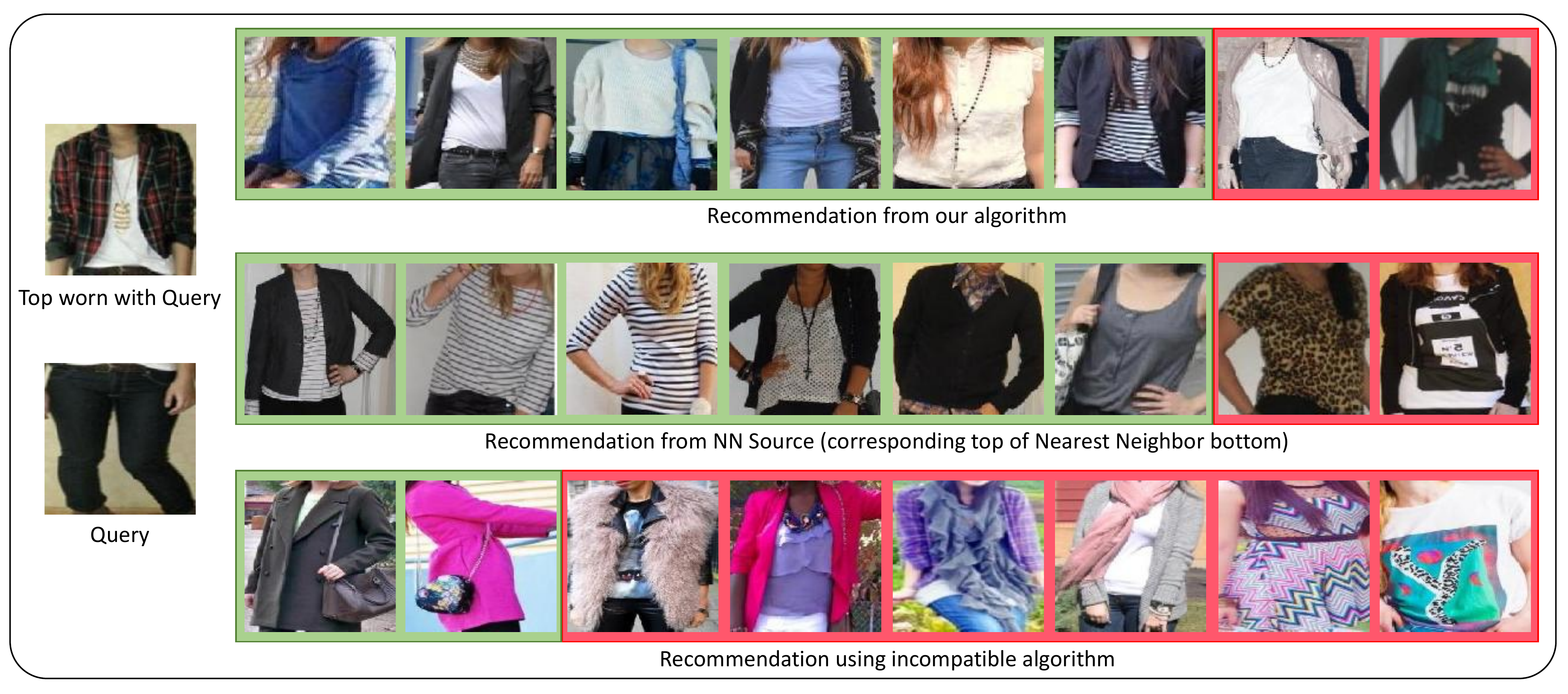}
		\label{fig:qualitative_1}
	}
	\subfigure[Complementary recommendation for a less common query item (pink skirt)]{
		\includegraphics[width=\textwidth,trim={0.2cm 0.3cm 0.2cm 0.3cm},clip]{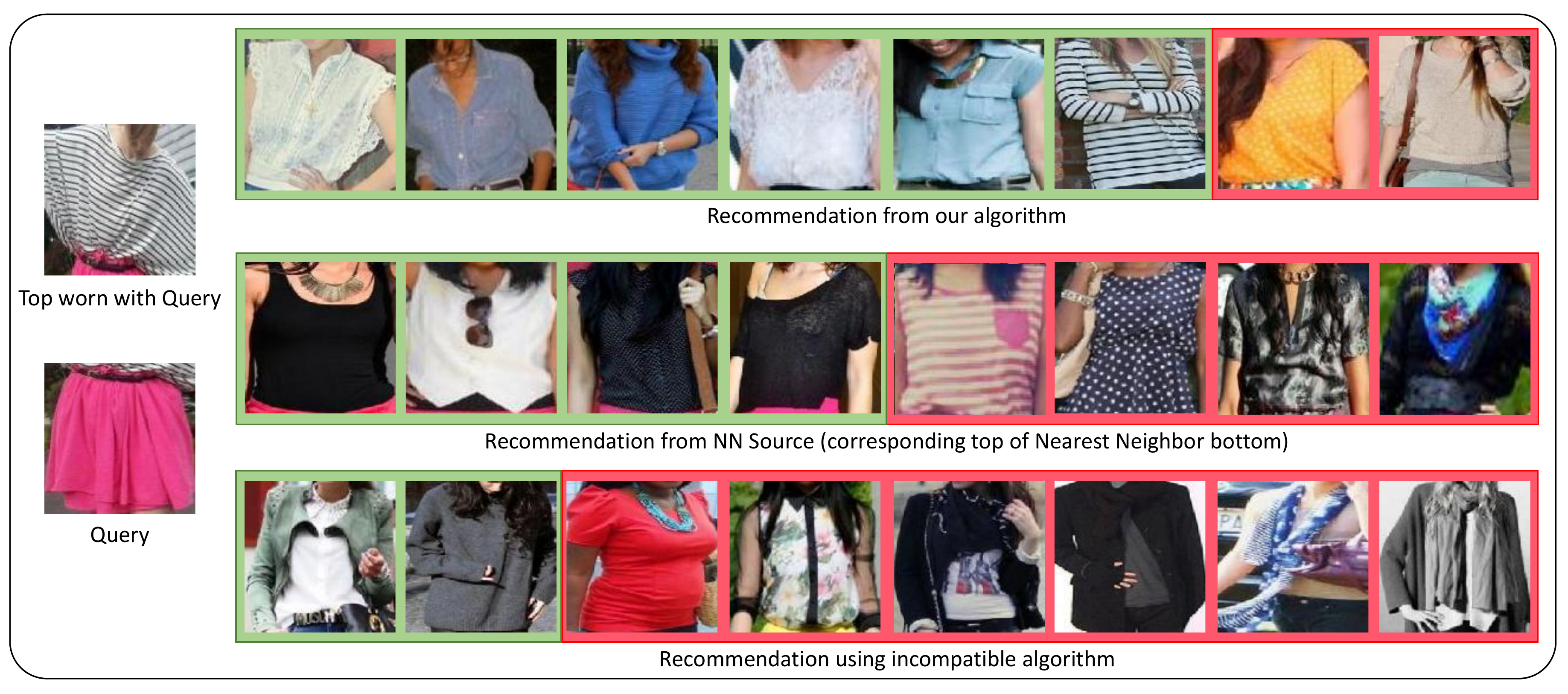}
		\label{fig:qualitative_2}
	}
	\caption{Comparison of the results from different recommendation algorithms. Highlighted in green are the items that have been marked as complementary to the query input by a fashion specialist. The CRAFT approach generates better (see Section \ref{subsect:user-study-analysis} for quantitative analysis) and diverse recommendations.}
	\label{fig:qualitative}
\vspace*{-3ex}
\end{figure*}

Figure~\ref{fig:qualitative} shows qualitative results of the different recommendation methods for two query items. For this experiment, we generated 8 top recommendations from each algorithm and asked a fashion specialist to identify the top items that complement the given bottom query. While all of the approaches produce visually diverse recommendations, not all of them are compatible with the query. 
For a common bottom outfit such as dark jeans (Figure~\ref{fig:qualitative_1}), NN-Source perform as well as CRAFT, while for a less common bottom such as bright pink skirt (Figure~\ref{fig:qualitative_2}) they perform worse (see Section~\ref{subsect:user-study-analysis} for a more thorough analysis). This is aligned with our intuition that the quality of NN-Source recommendation highly depends on the proximity of the neighbors of the query. Interestingly, the results of the `incompatible' algorithm demonstrate that our discriminator is able to learn not only the concept of visual compatibility, but also \emph{incompatibility}: it often produces unusual outfit recommendation (e.g., the fur top as the third item in Figure~\ref{fig:qualitative_1}) that is not likely to complement the given bottom query.

\subsection{User Study Design}
\label{subsect:user-study-design}

When recommendations are provided from an open ended set, they are difficult to evaluate in absolute terms. Typically, different recommendation approaches are compared via A/B testing. Furthermore, for subjective domains such as fashion, it is preferable to obtain input from domain experts who are familiar with nuances involved in making style-appropriate recommendations. We adopt A/B testing as the main methodology to compare our proposed approach to various baselines described in Section~\ref{subsect:baseline-algorithms}. We evaluate the relevance of recommendations generated by each algorithm by measuring their \emph{acceptance} by domain experts.

We approached a panel of four fashion specialists (FS) to provide feedback on recommendations generated by various algorithms. 
Each FS was presented with $17$ recommendations for a given query (bottom) item, for each of the four algorithms. Among these recommendations, the FS were asked to select those that they judge to be complementary to the query. 
We used a total of $64$ different query bottoms in this study, ranging for popular bottoms such as blue jeans to less common bottoms such as richly patterned skirts (see the Supplementary material for the full list). The images were presented to FS in a random order to eliminate any bias for the algorithm or query items. 

Since some FS are in general more selective than others, we need to normalize for their individual bias. To achieve this, we add the \textit{actual} top worn by the user in the query outfit to the set of $17$ recommendations at a random location. We normalize the FS acceptance scores by their likelihood of selecting the actual top as an acceptable recommendation. Note that we only perform analysis on the newly recommended tops, and exclude the original top from our results.


\subsection{Analysis}
\label{subsect:user-study-analysis}
\begin{figure}
	\begin{center}		
		\subfigure[Overall acceptance rates] {
			\includegraphics[width=\linewidth]{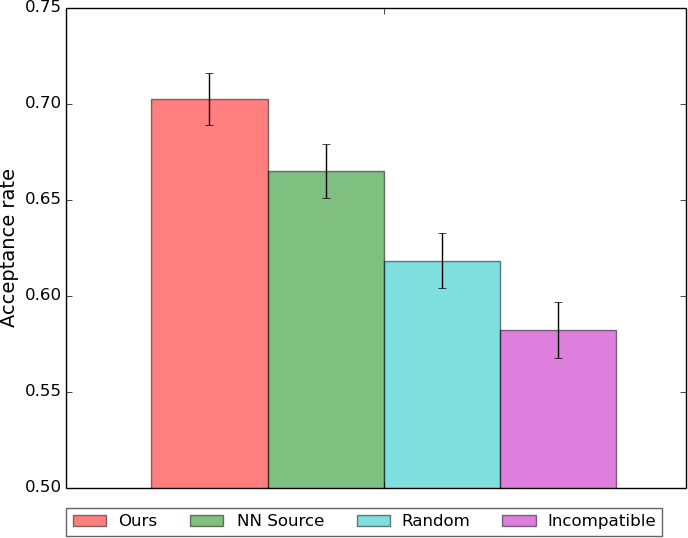}
			\label{fig:subjective_compatibility_1_bin}
		}
		\subfigure[Binned acceptance rates] {
			\includegraphics[width=\linewidth]{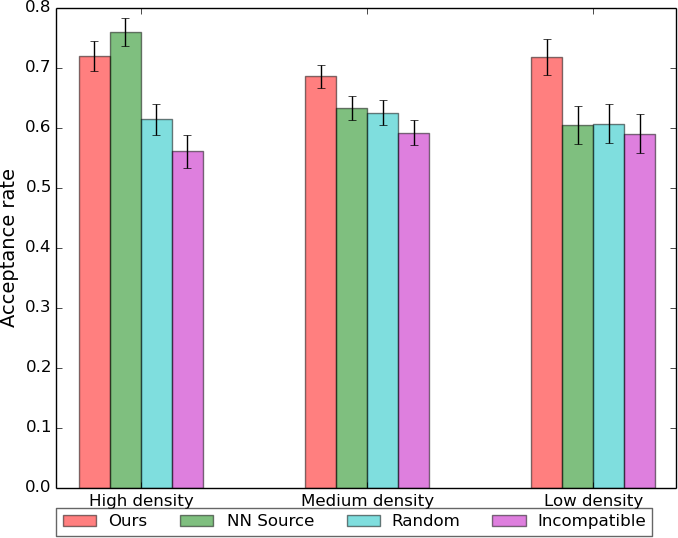}
			\label{fig:subjective_compatibility_3_bins}
		}
	\end{center}
	\caption{Mean acceptance rate of recommendations for the different algorithms, as evaluated by fashion specialists (error bars indicate $95$\% confidence intervals). (a) Overall acceptance rating for each algorithm. (b) Acceptance ratings binned according to the density (high, medium, low) of query items in the feature space.}
	\label{fig:compatibility_density}
\vspace*{-3ex}
\end{figure}
Figure~\ref{fig:subjective_compatibility_1_bin} shows the average rate of acceptance of generated recommendations for all FS for the four algorithms. As discussed, acceptance rates were normalized by the probability of each FS accepting the actual top for the given query bottom. The error bar denotes the $95\%$ confidence interval for each of the results. Non-overlapping error bars indicate that the differences between the two results are statistically significant. The NN-Source algorithm has the overall acceptance score of $66.5 \pm 1.4$ and works better than the \textit{Random} and \textit{Incompatible} baseline algorithms as expected. The CRAFT approach generates recommendations with the highest FS acceptance score ($70.3\pm 1.4$).

\textbf{Stratification by Feature Space Density:}
It is even more interesting to break down the analysis of the results in terms of the density of the query items in the feature space. Intuitively, the NN-Source algorithm should perform well in providing diverse complementary recommendations for high density regions in bottom feature space (\ie, popular bottoms such as blue jeans). However, a nearest neighbor based approach would not perform well in low density regions, corresponding to rare bottom examples (e.g., striped purple skirt).

To validate this hypothesis, we approximate the density of each query point by taking the average distance to $K=25$ nearest neighbors and bin the queries into low, medium, and high density regions, respectively. Figure~\ref{fig:subjective_compatibility_3_bins} shows the average recommendation acceptance rate provided by FS for each algorithm in each density region. Again, the error bars denote the $95\%$ confidence interval for each result. For queries that fall in the high density regions, the difference between our proposed approach and the NN-Source algorithm is statistically insignificant (error bars overlap). 
This is expected since nearest neighbor search is a good estimator of the conditional distribution of tops given a bottom for high density regions, where a large number of bottoms are available. However, the NN-Source algorithm starts to degrade at the medium density level, and eventually degenerates to similar performance as the \textit{Random} and the \textit{Incompatible} recommendation algorithms for low density regions.
In contrast, the performance of CRAFT is consistent across all regions and is better than baseline algorithms for mid and low density regime.
Thus, the proposed conditional transformer is able to generalize well irrespective of the density of the neighborhood surrounding the query item. 

\section{Conclusion and Future Work}
\label{sect:conclusion}
We presented CRAFT, an approach to visual complementary recommendation by learning the \emph{joint} distribution of co-occurring visual objects in an unsupervised manner. Our approach does not require annotations or labels to indicate complementary relationships. The feature transformer in CRAFT samples from a \emph{conditional} distribution to generate diverse and relevant item recommendations for a given query. The recommendations generated by CRAFT are preferred by the domain experts over those produced by competing approaches. 

By modeling the feature level distributions, our framework can potentially enable a host of applications, ranging from domain adaptation to one- or few-shot learning. The current work could be extended to incorporate the end-to-end learning of domain-related encoders as part of the generative framework.

{\small
\bibliographystyle{ieee}
\bibliography{references}
}

\end{document}